# Deep Doubly Supervised Transfer Network for Diagnosis of Breast Cancer with Imbalanced Ultrasound Imaging Modalities

Xiangmin Han[1], Jun Wang[1], Weijun Zhou[2], Cai Chang[3], Shihui Ying[4], Jun Shi[1(✉)]

[1] Shanghai Institute for Advanced Communication and Data Science, School of Communication and Information Engineering, Shanghai University, China
junshi@shu.edu.cn
[2] Department of ultrasound, The First Affiliated Hospital of Anhui Medical University, China
[3] Department of ultrasound, Fudan University Shanghai Cancer Center, Department of Oncology, Shanghai Medical College, Fudan University, China
[4] Department of Mathematics, School of Science, Shanghai University, Shanghai, China

**Abstract.** Elastography ultrasound (EUS) provides additional bio-mechanical information about lesion for B-mode ultrasound (BUS) in the diagnosis of breast cancers. However, joint utilization of both BUS and EUS is not popular due to the lack of EUS devices in rural hospitals, which arouses a novel modality imbalance problem in computer-aided diagnosis (CAD) for breast cancers. Current transfer learning (TL) pay little attention to this special issue of clinical modality imbalance, that is, the source domain (EUS modality) has fewer labeled samples than those in the target domain (BUS modality). Moreover, these TL methods cannot fully use the label information to explore the intrinsic relation between two modalities and then guide the promoted knowledge transfer. To this end, we propose a novel doubly supervised TL network (DDSTN) that integrates the Learning Using Privileged Information (LUPI) paradigm and the Maximum Mean Discrepancy (MMD) criterion into a unified deep TL framework. The proposed algorithm can not only make full use of the shared labels to effectively guide knowledge transfer by LUPI paradigm, but also perform additional supervised transfer between unpaired data. We further introduce the MMD criterion to enhance the knowledge transfer. The experimental results on the breast ultrasound dataset indicate that the proposed DDSTN outperforms all the compared state-of-the-art algorithms for the BUS-based CAD.

## 1 Introduction

B-mode ultrasound (BUS) is a clinical routine imaging tool to diagnose breast cancers. With the fast development of artificial intelligence technology, the BUS-based computer-aided diagnosis (CAD) has attracted considerable attention in recent years [1].

However, BUS only provides diagnostic information related to the lesion structure and internal echogenicity, which limits the performance of CAD to a certain extent.

Elastography ultrasound (EUS) imaging has emerged as an effective imaging technology for the diagnosis of breast cancers, which shows information pertaining to the biomechanical and functional properties of a lesion [2]. Joint utilization of both BUS and EUS provides complementary information for breast cancers to promote diagnostic accuracy [3]. However, the EUS devices are generally scarce in rural hospitals, which makes EUS not popular in diagnosing breast cancers in clinical practice.

Transfer learning (TL) aims to improve a learning model in the target domain by transferring knowledge from the related source domains [4][5]. TL has achieved great success in various classification tasks, including CAD [6][7]. Therefore, the performance of a single-modal imaging-based CAD model can be effectively promoted by transferring knowledge from other related imaging modalities or diseases [6].

It is worth noting that modality imbalance is a common phenomenon in clinical practice. That is, there are not only some paired BUS and EUS images with shared labels but also additional single-modal labeled BUS images in this work. Therefore, the source domain (EUS modality) has fewer samples than those in the target domain (BUS modality) in our work, which is contrary to the conventional TL applications. The inadequate data in the source domain also increase the difficulty for TL since it cannot provide enough supervision for TL. The conventional TL methods can handle this transfer task by performing the feature- or classifier-level transfer [4][6][8][9]. However, these TL methods have no constraints on the labels of both the source and target domains, and therefore cannot fully use the label information to explore the intrinsic relation between two modalities and then guide the promoted knowledge transfer.

Learning using privileged information (LUPI) is a newly proposed TL paradigm developed on the paired data in source and target domains with shared labels [10]. Support vector machine plus (SVM+) is a typical classifier under the LUPI paradigm, which generally outperforms the conventional TL classifiers due to the supervision of the shared labels [10]. However, SVM+ cannot conduct TL for unpaired or imbalanced data also due to the limitation of the LUPI paradigm.

On the other hand, convolutional neural network (CNN) based TL methods generally achieve superior performance to the conventional TL approaches in many classification tasks [11]. Although the source domain generally includes a large number of labeled data while the target domain only has a few labeled data, most of these works focus on the knowledge transfer between unpaired data in an unsupervised way[12].

Therefore, it is necessary to develop a new TL paradigm that can effectively address the issue of TL for imbalanced medical modalities in a supervised way. To this end, we propose a novel deep doubly supervised transfer network (DDSTN) for the BUS-based CAD of breast cancers. As shown in Fig. 1, this new TL paradigm doubly transfers knowledge between both the paired and unpaired data between the source and target domains in a unified framework. Specifically, the SVM+ classifier performs the transfer for the paired ultrasound data with shared labels, while the two-channel CNNs conduct another supervised transfer for the unpaired labeled data by MMD criterion. The double transfer mechanism can effectively adopt both shared and unshared labels to mine the intrinsic transferred information, and then guide the knowledge transfer from the limited samples in the source domain.

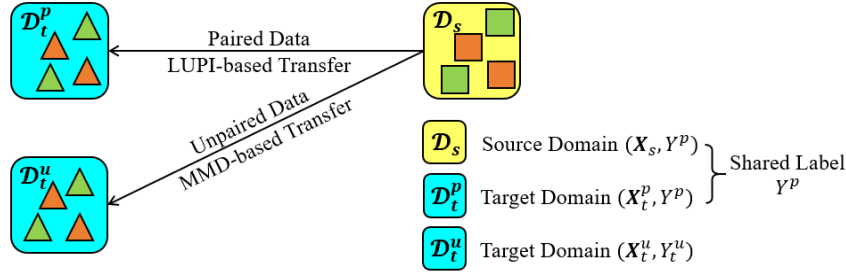

**Fig. 1.** Illustration of the proposed DDSTN paradigm. $(X_s, Y^p)$ and $(X_t^p, Y^p)$ denote the paired data with shared labels $Y^p$ in the source domain $\mathcal{D}_s$ and target domain $\mathcal{D}_t^p$, respectively. $(X_t^u, Y_t^u)$ is the additional single-modal data in the target domain $\mathcal{D}_t^u$.

The main contributions are twofold as follows:
1) We propose a new doubly supervised TL paradigm to address the issue of TL for imbalanced modalities with labeled data, which can not only make full use of the shared labels to effectively guide knowledge transfer, but also perform additional information transfer between unpaired data. Therefore, more transferred knowledge promotes the classification performance.
2) We develop a novel DDSTN algorithm to perform the doubly supervised TL from fewer EUS samples in the source domain to the BUS-based CAD for breast cancers. Specifically, DDSTN integrates the SVM+ paradigm for the TL of paired data and the deep TL network for transfer between the unpaired data into a unified framework. The experimental results show its effectiveness on the BUS-based CAD for breast cancers.

## 2 Method

### 2.1 Network Architecture of DDSTN

Fig. 2 shows the flowchart of our proposed DDSTN, which consists of two components, namely, LUPI-based supervised TL module for paired data and MMD-based supervised TL module for unpaired data. There are two independent CNNs for the source and target domains, respectively, which mainly learn feature representation, and also perform knowledge transfer for both paired and unpaired data.

In this work, BUS and EUS imaging work as target domain and source domain, respectively. We define $(X_s, Y^p)$ and $(X_t^p, Y^p)$ to be the paired data with shared labels $Y^p$ in the source domain $\mathcal{D}_s$ and target domain $\mathcal{D}_t^p$, respectively, and $(X_t^u, Y_t^u)$ is the additional single-modal data in the target domain $\mathcal{D}_t^u$. The superscript $p$ and $u$ denote paired and unpaired, the subscript $s$ and $t$ mean the source and target domain.

The LUPI-based supervised TL module performs knowledge transfer under the guidance of shared labels to promote the classifier in the target domain. As shown in Fig. 2, the loss function of LUPI contains coupled SVM+ loss. We optimize both the two-channel networks simultaneously with this coupled loss.

The MMD-based supervised TL module shares the same network with the LUPI-based supervised TL module. We integrate the MMD learning criterion and hinge loss into a uniform supervised architecture. MMD is used to minimize the distribution imparity between two domains, and the hinge loss in SVM can help to learn a strong classifier. Since we introduce the label information, the unpaired data can be trained in a supervised way. Moreover, it is worth noting that the hinge loss is just the same as the LUPI-based supervised TL module.

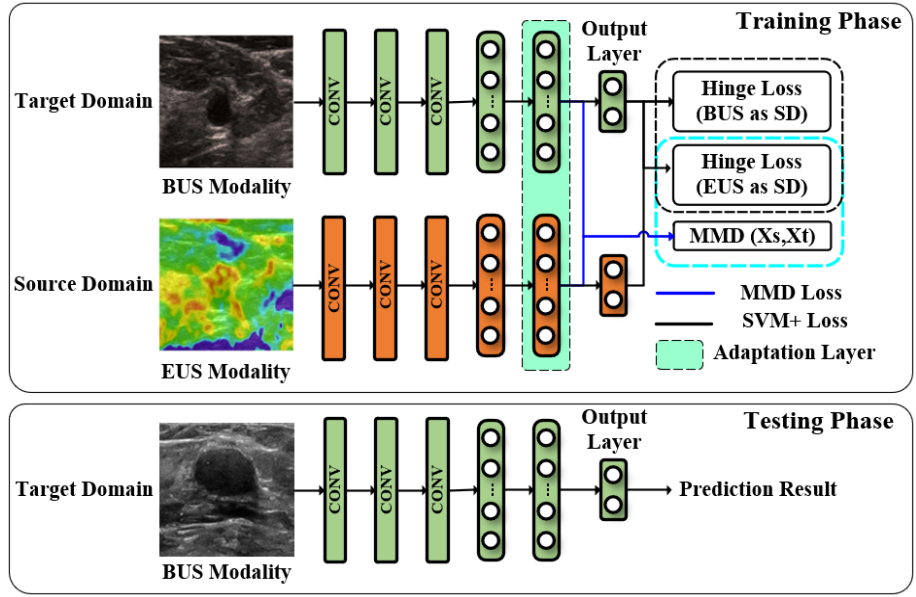

**Fig. 2.** The network architecture of proposed DDSTN. SD and TD denote the source domain and the target domain, respectively. The black dotted box represents the loss of LUPI, and the light green dotted box represents the supervised MMD criterion.

In the training phase, the source and target networks are optimized under an overall objective function, while in the testing phase, only the learned target networks (BUS modality network) is used to predict the results.

### 2.2 Doubly Supervised Transfer Learning

We propose a doubly supervised transfer strategy to perform knowledge transfer across the imbalance modality. The overall object function incorporates two loss parts for transferring the paired and unpaired data, respectively, into the following formula:

$$\mathcal{L} = \mathcal{L}_{paired} + \mathcal{L}_{unpaired} \tag{1}$$

where $\mathcal{L}_{paired}$ is the LUPI paradigm for the TL of paired data and $\mathcal{L}_{unpaired}$ is the MMD learning criterion for the TL of unpaired data.

**LUPI Paradigm for TL of Paired Data.** The LUPI paradigm is adopted to perform transfer for the paired data with shared labels [10]. Here, the typical SVM+ classifier is used with the objective function as following:

$$\mathcal{L}_{paired} = \min \frac{1}{2}(\|W_t\|^2 + \lambda_1 \|W_s\|^2) + C_1 \sum_{i=1}^{n^p}[\langle W_s, x_{s_i}^p \rangle + b_s] \tag{2}$$

$$\text{s.t. } y_i[\langle W_t, x_{t_i}^p \rangle + b_t] \geq 1 - [\langle W_s, x_{s_i}^p \rangle + b_s], \quad i = 1, \dots, n^p$$

$$\text{and } \langle W_s, x_{s_i}^p \rangle + b_s \geq 0, \quad i = 1, \dots, n^p$$

where $y_i \in Y^p$, $\{W_t, W_s\}$ and $\{b_t, b_s\}$ denote the weight matrices and bias vectors of the last layer in both target and source domains, respectively. $\lambda_1 > 0$ is a hyperparameter that restricts the correcting capacity, $C_1 > 0$ is a coefficient that balances the hinge loss term and the regularization term, $n^p$ is the number of paired data, and $\|\cdot\|$ denotes the L2-norm of the weight matrix.

As shown in Fig. 2, the LUPI paradigm for the paired data has a coupled loss for two domains. Thus, EUS and BUS modalities are alternately taken as the source domain data to perform TL to improve the network of the target domain.

**MMD Criterion for TL of Unpaired Data.** To conduct knowledge transfer between unpaired data, MMD is introduced to minimize the distribution imparity between two domains. By considering labels as the supervision to further improve the learning performance of the classifier, we design a new loss function for unpaired data:

$$\mathcal{L}_{unpaired} = \min(\frac{1}{2}\|W_t\|^2 + C_2 \sum_{j=1}^{n_t^u} \max(0, 1 - y_{t_j}^u [\langle W_t, x_{t_j}^u \rangle + b_t])$$

$$+ \lambda_2 \left\| \frac{1}{n_s^u} \sum_{k=1}^{n_s^u} \phi(x_{s_k}) - \frac{1}{n_t^u} \sum_{j=1}^{n_t^u} \phi(x_{t_j}^u) \right\|_{\mathcal{H}}) \tag{3}$$

where $x_{s_k} \in X_s$, $x_{t_j}^u \in X_t^u$, $y_{t_j}^u \in Y_t^u$, $\lambda_2$ is non-negative hyperparameter of MMD. $\phi(\cdot)$ is a feature mapping function, we aim to find an optimal $\phi(\cdot)$ that can train a robust classifier, $n_t^u, n_s^u$ are the number of BUS imaging and EUS imaging, respectively.

In order to minimize Eq. (3), we perform the domain adaption on the penultimate layer to transfer the knowledge from the source domain to the target domain for the unpaired features [15]. The supervised domain fusion loss makes the domains indistinguishable in the process of representation learning.

**Doubly Supervised TL Strategy.** The final objective function for doubly supervised TL is formulated by combining the $\mathcal{L}_{paired}$ and $\mathcal{L}_{unpaired}$ as following:

$$\mathcal{L} = \min \frac{1}{2}(\|W_t\|^2 + \lambda_1 \|W_s\|^2)$$

$$+ C_1 \sum_{i=1}^{n^p} \max(0, 1 - y_i[\langle W_s, x_{s_i}^p \rangle + b_t]) + C_2 \sum_{j=1}^{n_t^u} \max(0, 1 - y_{t_j}^u [\langle W_t, x_{t_j}^u \rangle + b_t])$$

$$+ \lambda_2 \left\| \frac{1}{n_s^u} \sum_{k=1}^{n_s^u} \phi(x_{s_k}) - \frac{1}{n_t^u} \sum_{j=1}^{n_t^u} \phi(x_{t_j}^u) \right\|_{\mathcal{H}} \quad (4)$$

where $C_1$ and $C_2$ are non-negative constants of LUPI paradigm and distance metric loss, respectively, $\lambda_1$ restrict the correcting capacity of the classifier, and $\lambda_2$ is non-negative hyperparameter of MMD.

The overall objective function is optimized by stochastic gradient descent [13]. As shown in Fig. 2, only the learned target domain network is used to predict the results. The objective function is given by:

$$\hat{Y} = WX_t + b \quad (5)$$

where $X_t \subset \{X_t^p, X_t^u\}$, $W$ and $b$ are the learned parameters in the training stage.

## 3 Experiments

### 3.1 Data Processing

We evaluated the proposed DDSTN algorithm on a bimodal breast ultrasound dataset sampled by one of the authors, in which 106 patients (54 benign tumors patients and 51 malignant cancer patients) have both BUS and EUS modalities, while the other 159 patients (81 benign tumors patients and 78 malignant cancer patients) only have BUS data. The approval from the ethics committee of the hospital was obtained, and all patients had signed informed consent.

The bimodal ultrasound images were acquired by the Mindray Resona7 ultrasound scanner with the L11-3 probe by an experienced sonologist. All the malignant cancers have been proved by the pathological diagnosis. A region of interest (ROI) including the lesion region was selected by an experienced sinologist from each ultrasound image. Noting that for the paired BUS and EUS images, only the ROI in BUS image was manually selected, and the same location of ROI was then automatically mapped to EUS imaging to obtain the ROI.

### 3.2 Experimental Setup

The proposed DDSTN was compared with the following related or state-of-the-art TL algorithms.
1) CNN-SVM: CNN-SVM is a single-channel CNN which is compared as a baseline, we selected ResNet18 as the classification network for single-modality BUS and replace the softmax classifier with SVM.
2) CNN-SVM+ [14]: CNN-SVM+ is another baseline which consists of two-channel CNNs and an SVM+ classifier. BUS is considered as the diagnostic modality, while EUS is the source domain.

3) DDC [15]: DDC is a typical deep TL algorithm which uses the MMD criterion as the distribution distance metric.
4) DAN [16]: Deep adaptation networks (DAN) is an improved DDC algorithm that replaces the MMD with multi-kernel MMD and then calculates the multiple layer losses.
5) Deep CORAL [17]: Deep correlation alignment (Deep CORAL) is a deep TL algorithm based on correlation alignment, which learns a second-order feature transformation to minimize the feature distance between the source and the target domain.

The 3-fold cross-validation was adopted to evaluate all the algorithms. Specifically, the 106 paired data were always fixed as training data for the LUPI-based TL module, and the 159 additional BUS data were divided into three groups. We selected two of three groups of additional BUS data and all the EUS images from the 106 paired data to form another training set for the MMD-based TL module, while the remaining one BUS group was set as testing data. The experiment repeated three times. The final results were presented with the format of the mean ± SD (standard deviation).

The commonly used classification accuracy (ACC), sensitivity (SEN), specificity (SPE) and Youden index (YI) were selected evaluation indices. Moreover, the receiver operating characteristic (ROC) curve and the area under ROC curve (AUC) were also adopted for evaluation.

### 3.3 Experimental Results

Table 1 shows the classification results of different algorithms. It can be found that the proposed DDSTN outperforms all the compared algorithms with the best accuracy of 86.79±1.54%, sensitivity of 86.45±1.44%, specificity of 87.31±4.37%, and YI of 73.77±3.17%. DDSTN improves at least 1.92%, 2.04%, 0.4%, and 3.85% on accuracy, sensitivity, specificity and YI, respectively compared with other algorithms.

**Table 1.** Classification results of different algorithms

|  | ACC (%) | SEN (%) | SPE (%) | YI (%) |
|---|---|---|---|---|
| CNN-SVM | 82.34±5.67 | 81.56±4.22 | 84.56±1.22 | 66.12±3.51 |
| CNN-SVM+ | 84.87±2.85 | 84.41±4.45 | 85.28±1.41 | 69.69±5.63 |
| DDC | 83.33±1.44 | 81.53±3.02 | 85.40±4.68 | 66.93±2.97 |
| DAN | 84.85±1.11 | 83.01±2.61 | 86.91±3.58 | 69.92±2.27 |
| Deep CORAL | 84.47±2.31 | 84.07±3.12 | 85.66±2.14 | 69.73±2.25 |
| **DDSTN (proposed)** | **86.79±1.54** | **86.45±1.44** | **87.31±4.37** | **73.77±3.17** |

The experiments show that CNN-SVM+ achieves superior performance to CNN-SVM, which indicates the effectiveness of transferring information from EUS for the BUS-based CAD by LUPI paradigm. It also can be found that DDSTN improves at least 1.94% on accuracy, 2.38% on sensitivity, 0.40% on specificity and 3.85% on YI compared with DDC, DAN and Deep CORAL, which indicates the effectiveness of our doubly supervised TL paradigm. Moreover, DDSTN improves 1.92%, 2.04%, 2.03%, and 4.08% on accuracy, sensitivity, specificity and YI, respectively, over CNN-SVM+, suggesting the positive effect of TL between unpaired data for learning an effective classifier.

Fig. 3 shows the ROC curves and the corresponding AUC values of different algorithms. DDSTN again achieves the best AUC value of 0.871, which improves at least 0.028 over all the other algorithms.

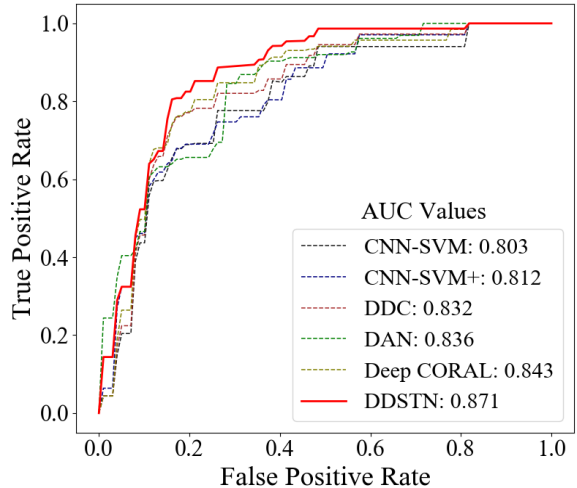

**Fig. 3.** ROC curves of different algorithms with the corresponding AUC values.

## 4   Conclusion

In summary, we propose a novel doubly supervised TL paradigm to address the issue of TL between imbalanced modalities with labeled data. The proposed DDSTN algorithm effectively performs the double supervised transfer between both the paired data with shared labels by the SVM+ paradigm and the unpaired data with different labels by the MMD criterion in a unified framework. The experimental results indicate that DDSTN outperforms all the compared algorithms on the BUS-based CAD for breast cancers.

In current work, we adopt MMD as the distribution distance metric for TL, and therefore we select DDC, DAN and Deep CORAL for comparison, since all these algorithms are developed based on MMD or MMD related criterion. In our future work, we will further improve the doubly supervised transfer network by studying other TL methods instead of MMD. Moreover, we will try to integrate the advantages of adversarial domain adaption networks in this new doubly supervised TL paradigm.

**Acknowledgements.** This work is supported by the National Natural Science Foundation of China (81830058, 81627804), the Shanghai Science and Technology Foundation (17411953400, 18010500600), the 111 Project (D20031) and the Nanjing Science and Technology Commission (201803027).